\documentclass[12pt,twocolumn,twoside]{article}

\usepackage{enumitem}
\usepackage{hyperref}
\usepackage{booktabs}
\usepackage[normalem]{ulem}
\usepackage{graphicx}
\useunder{\uline}{\ul}{}
\makeatletter
\newcommand\footnoteref[1]{\protected@xdef\@thefnmark{\ref{#1}}\@footnotemark}
\makeatother
\newcommand{\etal} {{et al.\/} }
\usepackage[para]{threeparttablex}

\usepackage{multirow}
\usepackage{multicol}
\usepackage{wrapfig}
\usepackage{titlesec}
\usepackage{blkarray}
\usepackage{fancyhdr}
\usepackage[numbers,comma]{natbib}
\usepackage{graphicx, caption}
\usepackage{subfig}
\usepackage{enumitem,booktabs}
\usepackage{rotating}
\usepackage{amsmath, amssymb, amsthm}
\usepackage{algorithm}
\usepackage[noend]{algpseudocode}
\usepackage{tikz}
\usepackage{adjustbox}
\usepackage{multirow}
\usetikzlibrary{patterns}
\usepackage{pgfplots}
\usetikzlibrary{pgfplots.groupplots}
\pgfplotsset{compat=1.12}
\tikzset{
    axis break gap/.initial=1mm
}
\pgfplotsset{ every non boxed x axis/.append style={x axis line style=-},
     every non boxed y axis/.append style={y axis line style=-}}
\usepackage{multirow}
\usepackage{siunitx}
\usepackage{booktabs, longtable}
\usepackage[top=1in,bottom=1in,left=1in,right=1in]{geometry}
\usepackage{fancyhdr}
\usepackage{hyperref}
\usepackage{color}
\usepackage{setspace,verbatim}
\usepackage{authblk}
\usepackage{transparent}

\usepackage{xcolor,colortbl}
\usepackage{url}
\usepackage{hyperref}

\usepackage{graphicx}

\definecolor{Gray}{gray}{0.85}
\definecolor{LightCyan}{rgb}{0.88,1,1}
\usepackage{array}
\newcolumntype{a}{>{\columncolor{Gray}}c}
\newcolumntype{b}{>{\columncolor{white}}c}
\newcolumntype{C}[1]{>{\centering\let\newline\\\arraybackslash\hspace{0pt}}m{#1}}
\newcolumntype{L}[1]{>{\raggedright\let\newline\\\arraybackslash\hspace{0pt}}m{#1}}

\usepackage[customcolors]{hf-tikz}

\tikzset{style green/.style={
    set fill color=green!50!lime!60,
    set border color=white,
  },
  style cyan/.style={
    set fill color=cyan!90!blue!60,
    set border color=white,
  },
  style orange/.style={
    set fill color=orange!80!red!60,
    set border color=white,
  },
  hor/.style={
    above left offset={-0.15,0.31},
    below right offset={0.15,-0.125},
    #1
  },
  ver/.style={
    above left offset={-0.1,0.3},
    below right offset={0.15,-0.15},
    #1
  }
}

\titleformat*{\section}{\normalsize\bfseries\sffamily}
\titleformat*{\subsection}{\small\bfseries\sffamily}
\titleformat*{\subsubsection}{\small\bfseries\sffamily}

\usepackage[font=small,labelfont=bf]{caption}

\newcommand\shorttitle{Discussion on Practical Sparse Regression}
\newcommand\authors{Sarwar, Sauk, and Sahinidis}

\fancypagestyle{firststyle}
{
   \fancyhf{}

   \fancyfoot[L]{\footnotesize\scshape \copyright 2020. The formal publication of this article is in {\em Statistical Science,} 35, 593--601, 2020 and available at \url{http://dx.doi.org/10.1214/20-STS806}.}
}

\title{\textbf{A Discussion on Practical Considerations with Sparse Regression Methodologies}
}
\author[1]{Owais~Sarwar}
\author[1]{Benjamin~Sauk}
\author[2]{Nikolaos~V.~Sahinidis}
\affil[1]{Carnegie Mellon University, Pittsburgh, PA, USA}
\affil[2]{Georgia Institute of Technology, GA, USA}
\date{}

\begin{document}

\begin{singlespacing}
\twocolumn[{%
\maketitle
\thispagestyle{firststyle}



\footnotesize
\vspace{-1cm}
\begin{abstract}
Sparse linear regression is a vast field and there are many different algorithms available to build models. Two new papers published in \textit{Statistical Science} study the comparative performance of several sparse regression methodologies, including the lasso and subset selection. Comprehensive empirical analyses allow the researchers to demonstrate the relative merits of each estimator and provide guidance to practitioners. In this discussion, we summarize and compare the two studies and we examine points of agreement and divergence, aiming to provide clarity and value to users. The authors have started a highly constructive dialogue, our goal is to continue it. 
\end{abstract}

\vspace*{0.8cm}

}]

\footnotesize

\section*{Introduction}
Research in sparse linear regression has covered considerable ground in recent decades. Trevor Hastie, Robert Tibshirani, and Ryan Tibshirani's \textit{Best Subset, Forward Stepwise, or Lasso? Analysis and Recommendations Based on Extensive Comparisons} and Dimitri Bertsimas, Jean Pauphilet, Bart van Parys' \textit{Sparse Regression: Scalable Algorithms and Empirical Performance} are important additions to the conversations in linear model selection.

Over two centuries have passed since Gauss and Legendre laid the foundations for the Ordinary Least Squares (OLS) method that is central to linear regression \cite{Stigler1988}. Since then, the success of OLS in the field has been tempered only by its limitations when the amount of regression variables becomes large. When there are many variables, OLS leads to over-fit models with poor accuracy and poor interpretability. 
As a consequence, statisticians have concentrated their attention on methods that build models with only a small subset of the total regression variables--i.e., sparse estimators \cite{Hastie2015}. 

In the 1960s, Hocking and Leslie~\cite{Hocking1967} wrote about using forward selection to select from a few dozen regression variables, not a challenging task even for the computers of the time. Finding an optimal subset however was much harder---Furnival and Wilson's 1974 \textit{Leaps and Bounds} algorithm struggled with these size problems \cite{Furnival1974}. Following the work of Hoel and Kennard \cite{Hoerl1970}, ridge regression became popular in the 70s and 80s for reducing the variance of the OLS estimators, but unfortunately doing so without inducing sparsity. In the 1990s, work by Breiman on the nonnegative garrote \cite{Breiman1995} ultimately inspired Tibshirani's sparse, efficient, and ultra-popular lasso methodology \cite{Tibshirani1996}. The lasso allows for simultaneous coefficient shrinkage to reduce variance, and variable selection by setting many coefficients to zero. In the quarter century since, numerous refinements of the lasso have been proposed \cite{Meinshausen2007, Zou2005, Zou2006, Wang2011}, nonconvex penalties were developed \cite{Fan2001, Zhang2010, Mazumder2011}, and other alternatives \cite{Candes2007, Meinshausen2010} including, recently, best subset selection \cite{Cozad2014, Bertsimas2015} have gained attention. Today, regression problems with millions of variables are within the reach of the average user. 

These established regression methods largely come with a body of theoretical and experimental analyses to guide development and practice. Nevertheless, most of these analyses are narrow and few studies extensively compare the empirical performance of estimators at a macro-level (an exception, \cite{Wang2020}). As Bertsimas~\etal note, ``the profusion of research ... might have caused confusion and provided little guidance to practitioners."
We compare how these two recent studies endeavour to deliver us from this incertitude. Although Bertsimas~\etal also examine classification, we will only focus on regression. 

Before discussing the papers, we commend both groups for making the code for their research available open-source. Hastie~\etal go notably further by distributing the code for not only their regression methods, but also for the scripts used to run their actual simulations. 

\section*{Comparing Regression Methods}
Both papers compare the performance of multiple popular regression algorithms on synthetic data generated according to a linear model with variable constant Gaussian error in the response. While both sets of authors consider firmly-entrenched $l_1$-based regression methods against their $l_0$-based (or generally nonconvex) alternatives, the conclusions they reach are not completely aligned. 
Although these studies share some similarities, in this section we examine some substantial differences between them. While a detailed accounting here is not practical, we highlight the difference in the estimators compared, scale of the problems considered, and the differing nature of experiments conducted. A thorough summary of the experimental setups is provided for reference in Table \ref{Tab: Comparison}, in the caption of which we explain notation that will be referenced later. 

We also discuss some conclusions and perspectives on regression articulated in the two papers, including on the predictive power of the various methods, the sparsity advantage of subset selection and the other lasso-alternatives, and the practical philosophy behind choosing between regression methods. A broader summary of the authors' main points is provided for convenience in Table \ref{Tab: mainpoints}.

For our computations, we use the \texttt{ncvreg} package,\footnote{\label{ncvregnote}\href{https://cran.r-project.org/web/packages/ncvreg/index.html}{R Package Link}} 
\texttt{L0Learn} package,\footnote{\label{l0learnref}\href{https://cran.r-project.org/web/packages/L0Learn/index.html}{R Package Link}}
and code provided by the authors.\footnote{\label{bestsubsetref}\href{https://github.com/ryantibs/best-subset/}{R Package Link}}$^{,}$
\footnote{\label{cioref}\href{https://github.com/jeanpauphilet/SubsetSelectionCIO.jl}{Julia Package}}$^{,}$
\footnote{\label{ssref}\href{https://github.com/jeanpauphilet/SubsetSelection.jl}{Julia Package Link}} Our scripts are available online.\footnote{\href{https://github.com/osarwar/stsdiscussion2020}{Scripts Link}}

\begin{table*}[p!]
\caption{Summary of comparison frameworks for both papers. We retain the notation of the two papers discussed; here, $p$, $n$, $k_{true}$, and $SNR$ refer to, respectively, the total number of regressors considered, the number of data points, the true number of regressors, and the signal-to-noise ratio.}
\label{Tab: Comparison}
\centering
\begin{adjustbox}{width=0.84\textwidth}
\begin{tabular}{@{}cll@{}}
\toprule
\multicolumn{1}{c|}{Feature}                                                     & Hastie~\etal                                                                                                                                                                                                  & Bertsimas~\etal                                                                                                                                                                                                                                                                                                                                                                                                      \\ \midrule
\multicolumn{1}{c|}{Estimators}                                                    & \begin{tabular}[c]{@{}l@{}}Lasso \cite{Tibshirani1996}\\ Relaxed Lasso (rlasso) \cite{Meinshausen2007}\\ Forward Stepwise Selection \\ (FS) \cite{Hocking1967}\\ Best Subset MIO (BSS) \cite{Bertsimas2015} \\ (cardinality-constrained)\\ \textit{In supplement:} \\  L0Learn \cite{Hazimeh2018}\\ (unregularized and \\ $l_1$-regularized)\\ SparseNet \cite{Mazumder2011}\end{tabular} & \begin{tabular}[c]{@{}l@{}} Lasso/ENet \cite{Zou2005}\\ MCP \cite{Zhang2010}\\ SCAD \cite{Fan2001}\\ Best Subset CIO (CIO) \cite{Bertsimas}\\ (cardinality-constrained, \\ $l_2$-regularized)\\ Boolean Relaxation of Best Subset \\ (SS) \cite{Pilanci2015}\\ (cardinality-constrained, \\ $l_2$-regularized)\end{tabular}                                                                                                                                                                                                         \\ \cmidrule(l){2-3} 
\multicolumn{1}{c|}{\textit{p}}     & \begin{tabular}[c]{@{}l@{}}                       10, 100, $10^3$ \\ $ n > p$ and $p > n$                  \end{tabular}                                                                                                                                                                                  &  \begin{tabular}[c]{@{}l@{}} $2\times10^4, 10^4, 2\times10^3$ \\  $p > n$            \end{tabular}                                                                                                                                                                                                                                                                                                                                                                                            \\ \cmidrule(l){2-3} 
\multicolumn{1}{c|}{\textit{n}}                                                    & \begin{tabular}[c]{@{}l@{}}50 ($p=10^3$)\\ 100 ($p=\{10, 10^3\}$)\\  500 ($p=10^2$)\end{tabular}                                                                                                                          & 500--thousands                                                                                                                                                                                                                                                                                                                                                                                                       \\ \cmidrule(l){2-3} 
\multicolumn{1}{c|}{$k_{true}$}                                              & \begin{tabular}[c]{@{}l@{}}5 ($p=\{10, 10^2, 10^3\}$)\\ 10 ($p=10^3$)\end{tabular}                                                                                                                                      & \begin{tabular}[c]{@{}l@{}}10 ($p=2\times10^4$, $SNR=0.05$)\\ 50 ($p=10^4$, $SNR=1$)\\ 100 ($p=2\times10^4$, $SNR=6$)\end{tabular}                                                                                                                                                                                                                                                                                                                   \\ \cmidrule(l){2-3} 
\multicolumn{1}{c|}{\begin{tabular}[c]{@{}c@{}}Sparsity \\ structure\end{tabular}} & \begin{tabular}[c]{@{}l@{}}4 patterns \\ 1 considers weak sparsity\end{tabular}                                                                                                                                & Random                                                                                                                                                                                                                                                                                                                                                                                                                \\ \cmidrule(l){2-3} 
\multicolumn{1}{c|}{Correlation}                                                   & \begin{tabular}[c]{@{}l@{}}Toeplitz: predictor covariance \\ $\Sigma_{i,j} = \rho^{|i-j|}$\\ $\rho = \{0, 0.3, 0.7\}$\end{tabular}                                                                                                        & \begin{tabular}[c]{@{}l@{}}Toeplitz, $\rho = \{0.2, 0.7\}$\\ (Mutual Incoherence Condition (MIC) \\true)\\``Hard" structure (MIC fails)\cite{Loh2017}\end{tabular}                                                                                                                                                                                                                                                                                                       \\ \cmidrule(l){2-3} 
\multicolumn{1}{c|}{\textit{SNR}}                                                  & 0.05--6 (logarithmic scale)                                                                                                                                                                                   & 0.05, 1, 6                                                                                                                                                                                                                                                                                                                                                                                                             \\ \cmidrule(l){2-3} 
\multicolumn{1}{c|}{Metrics}                                                       & \begin{tabular}[c]{@{}l@{}}Relative (to Bayes) Test Error\\ Proportion of Variance \\ Explained (PVE) \\ Number Nonzeros\\ F-score (harmonic avg. of \\precision and recall)\end{tabular}                              & \begin{tabular}[c]{@{}l@{}}Accuracy\\ (\% true regressors recovered) \\ False Discovery Rate (FDR)\\Test Mean Squared Error (MSE)\end{tabular}                                                                                                                                                                                                                                                                        \\ \cmidrule(l){2-3} 
\multicolumn{1}{c|}{Experiments}                                                   & \begin{tabular}[c]{@{}l@{}}Evaluated test metrics across \\ range of data settings after \\ fitting each estimator \\ using cross validation.\end{tabular}                                                                                                                & \begin{tabular}[c]{@{}l@{}}Looked at both correlation structures.\\ When MIC holds, looked at high \\ and low correlation level. \\ \\ Fixed support size $k = k_{true}$; \\ evaluated Accuracy, FDR, and \\ MSE for 3 $(p, k, SNR)$ settings \\ with asymptotically increasing $n$. \\ \\ Cross-validated support size $k$; \\ evaluated MSE with fixed \\ $(p, k, SNR, n)$; evaluated Accuracy and \\ FDR with for 3 $(p, k, SNR)$ settings \\  with asymptotically increasing $n$.\end{tabular} \\ \bottomrule
\end{tabular}
\end{adjustbox}
\end{table*}

\begin{table*}[p!]

\caption{Summary of major conclusions from both studies. Clarifying notes added parenthetically. See caption for Table \ref{Tab: Comparison} for notation.}
\label{Tab: mainpoints}
\centering
\begin{adjustbox}{width=0.9\textwidth}
\begin{tabular}{c|ll}
\toprule
                                                      Method                   & Hastie~\etal                                                                                                                                                                                                                                                                                                                                                                                                                                                                                                                                                                                              & Bertsimas~\etal                                                                                                                                                                                                                                                                                                                                                                                                                                                                                                                                                                                                                                                                  \\ \hline
\multicolumn{1}{c|}{Lasso}                                                      & \begin{tabular}[c]{@{}l@{}}Lasso more accurate (in terms of\\ error) in low SNR range, \\ best subset more accurate in \\ high SNR.\\ \\ (In terms of variable selection, \\ when $p > n$ the lasso is inferior\\ to subset methods.)\end{tabular}                                                                                                                                                                                                                                                                                                                                                          & \begin{tabular}[c]{@{}l@{}}Lasso has the lowest Accuracy\\ and highest FDR of all methods \\ across settings. When MIC does \\ not hold, this difference is large.\\ When noise is large, lasso \\competes with other methods. \\ \\ (Above conclusions mainly from\\ setting where $k =k_{true}$. When \\ support is cross-validated, Lasso is \\ most Accurate but highest FDR.) \end{tabular}                                                                                                                                                                                                                                                                           \\
\multicolumn{1}{c|}{\begin{tabular}[c]{@{}l@{}}Subset\\ Selection\end{tabular}} & \begin{tabular}[c]{@{}l@{}}Best Subset Selection does not \\ perform much better than Forward\\ Selection, and is much less tractable.\\ \\ L0Learn1 (unregularized) and L0Learn2\\ ($l_1$-regularized) do not improve \\ much over BSS/FS. (In our view, \\L0Learn2 is better than L0Learn1, \\ and 
does improve over other subset \\ methods.)\end{tabular}                                                                                                                                                                                                                                                                                                                                                  & \begin{tabular}[c]{@{}l@{}}\\Both subset selection estimators \\ more Accurate (at $k =k_{true}$)\\ and have a much lower FDR \\ than alternatives. SS heuristic  \\ almost as good as optimal \\ algorithm. \\ \\ $l_2$-regularization greatly improves\\performance in noisy settings. \\ \\ Still expensive, but affordable for \\ many problems.\end{tabular}                                                                                                                                                                                                                                                                                                       \\
\multicolumn{1}{c|}{Alternatives}                                               & \begin{tabular}[c]{@{}l@{}}\\The relaxed lasso generally \\ outperforms all other methods. \\ \\ Nonconvex SparseNet is also good\\ overall but not quite as good as relaxed lasso.\end{tabular}                                                                                                                                                                                                                                                                                                                                                                                                         & \begin{tabular}[c]{@{}l@{}} \\ Nonconvex penaltizers (esp. MCP)\\ preferable to the lasso in terms of\\ test error and variable selection but \\ MCP has higher FDR than \\ subset methods.\end{tabular}                                                                                                                                                                                                                                                                                                                                                                                                                                                                 \\
\multicolumn{1}{c|}{Meta}                                                       & \begin{tabular}[c]{@{}l@{}}\\ \\ BSS not inherent `holy grail' of \\ regression. Different methods consider \\ different sets of bias-variance trade-offs \\ (depending on problem). Favored \\ set depends on problem class.\\ \\ Regression problems are very noisy. \\ Practical simulations consider ``low"\\ SNR values that faithfully capture  \\range of maximal PVE achievable \\ by linear regression model.\\ \\ Ultimately, all methods similar in \\PVE. This is meaningful because \\ PVE is a critical, intuitive \\ metric. Therefore, the most \\ convenient methods should be \\ preferred.\end{tabular} & \begin{tabular}[c]{@{}l@{}}\\ \\ Lasso popular due to performance, \\but also efficiency and accessibility. \\ \\Robustness and sparsity are different \\ but equally important objectives.\\ Ideal estimators combine robustness\\ of convex penalizers with sparsity \\ of nonconvex methods. \\ \\ Variable selection (sparsity) is \\ critical and FDR is as important as \\ Accuracy. Higher compute times \\ can be justified for more \\ interpretable models. \\ \\ In practice, subset selection \\ is limited in some time-sensitive \\ applications. Lasso/ENet could be \\ used for dimensionality reduction \\ followed by subset methods.\end{tabular} \\ \bottomrule 

\end{tabular}
\end{adjustbox}
\end{table*}

\subsection*{\textit{Extensive Comparisons} and \textit{Sparse Regression}, side-by-side}
It is important to understand how these studies differ to appreciate and contextualize their conclusions, and to find guidance for further research. 

\subsubsection*{Estimators Considered, Convex and Noncovex} 

Both studies highlight $l_1$/lasso-based and subset selection regression. While Hastie~\etal study the ordinary lasso, Bertsimas~\etal choose to use the related elastic net (ENet) \cite{Zou2005} which, theoretically, should improve upon the lasso's performance in high-dimensional, highly-correlated data while maintaining a similar (if slightly lower) level of sparsity. While there are differences between the two, they are likely not significant enough to change the authors' conclusions about the comparative performance of $l_1$ regression and subset selection. 

Both groups of authors consider both optimal and approximate versions of the cardinality-constrained version of subset selection. Hastie~\etal look at the unregularized versions (i.e. with no shrinkage penalties) whereas the optimal combinatorial integer optimization (CIO) method of Bertsimas~\etal \cite{Bertsimas} and the corresponding Boolean relaxation (SS) \cite{Pilanci2015} approximation, introduce an $l_2$-regularization penalty in the objective function that should improve the performance in the high-noise regime. Subset selection will be discussed further later. 

\begin{figure*}[htb!]
\center{\includegraphics[width=\textwidth]
{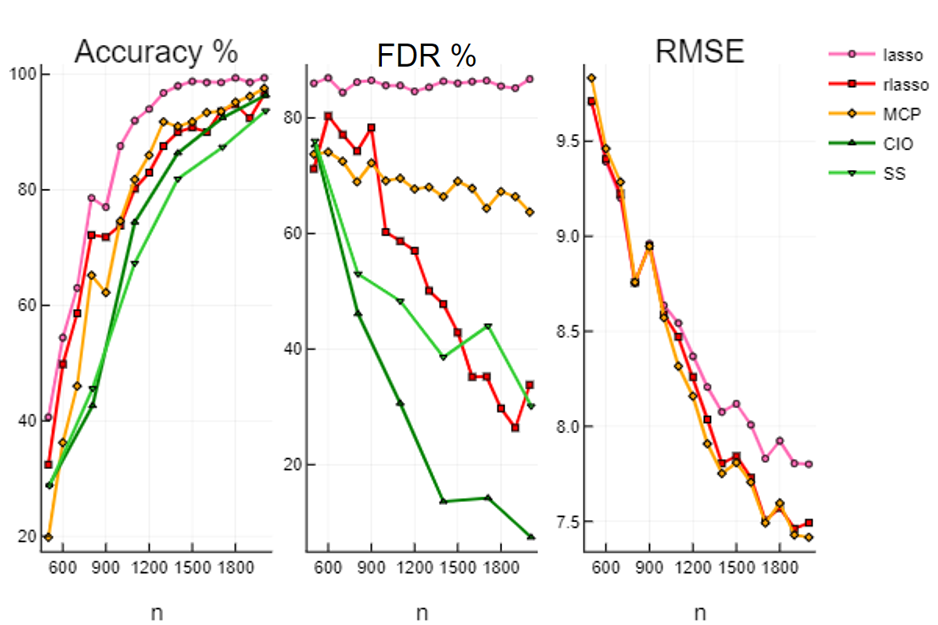}} 
\caption{\label{fig:rlassocomp} Relaxed lasso performance compared with the lasso and nonconvex alternatives.  For lasso and MCP, we used the default options from the R implementations. For rlasso, the $nlambda$ option was set to 50, with 10 interpolation points each. Data for CIO/SS is taken from Bertsimas~\etal ``Accuracy \%" refers to the percentage of the true regressors identified, ``FDR \%" refers to the false discovery rate (i.e., the percentage of regressors chosen that are erroneously included), and ``RMSE" refers to the root mean square prediction error on a test set of data size $n$.}
\end{figure*}

Each paper also considers alternatives. Hastie~\etal examine the relaxed lasso (rlasso) while Bertsimas~\etal discuss prominent nonconvex-estimators MCP \cite{Zhang2010} and SCAD \cite{Fan2001}. Bertsimas~\etal demonstrate that the lasso, in general, selects more irrelevant variables than the alternatives. As the authors note, this is not surprising because a wide-body of theoretical and empirical results spanning the past two-decades has demonstrated the limitations of the lasso---and of convex penalties in general---in high dimensions even in `friendly' designs (e.g., recently, \cite{Celentano2019}). We see this, too, in Hastie~\etal where the lasso is by far the least sparse method. These limitations have prompted the development of the MCP/SCAD (along with other nonconvex regularizers, e.g., bridge regression \cite{Frank1993}) and lasso-modifications such as the relaxed lasso. Bertsimas~\etal conclude that the ``best approaches ... combine a convex and nonconvex component." 
While that statement is sufficiently general as to admit multiple interpretations, it seems to be more intuitively the case for approaches where regression is done in a single stage. 
Consider that a two-stage approach that does not require minimization of a nonconvex function, the relaxed lasso (which performs variable selection and shrinkage in two separate steps), was the ``overall winner" in the studies by Hastie~\etal  It would be informative to analyze its performance within the framework of Bertsimas \etal To this end, we reproduce the ``Medium noise, High correlation" setting in Bertsimas~\etal\footnote{\label{expnote}Where $(p, k_{true}, SNR, \rho, MIC)=(10^4, 50, 1, 0.7, true)$. Validation is performed on a separate dataset of size $n$. For practical reasons, we use data directly from \textit{Sparse Regression} Figs. 6-7 to plot CIO and SS Accuracy/FDR. Results are averaged over 10 trials.} and plot results in Figure \ref{fig:rlassocomp}. These experiments were actualized because the authors have provided open-source implementations.

While our results for MCP are slightly different than those from Figs. 6-7 in Bertsimas~\etal (in particular, the FDR is worse here for large $n$; likely there is an implementation difference), even compared to their data the relaxed lasso appears to be superior to MCP in this setting. The cardinality-constrained estimators CIO/SS have a lower FDR, but the relaxed lasso is competitive with SS and has greater Accuracy than both. 

\subsubsection*{Experimental Design}

Bertsimas~\etal devote considerable effort to simulations that fix the size of the support to equal the size of the true support and evaluate the relative asymptotic performance of the estimators as $n$ is increased. While this approach is important to mirror the set-up and conclusions of important theoretical work on the lasso, it somewhat exaggerates the lasso's relative deficiencies by implying that the number of true positives the lasso will identify (which they quantify using ``Accuracy") is much lower than in actuality. 

In their own words, the authors of \textit{Sparse Regression} intend to analyze the estimators ``with an eye towards practicality." In our view, they achieve this goal more directly in simulations where the support is chosen by cross validation. Here, Bertsimas~\etal show that the lasso recovers more of the true support than the other methods (even when the Mutual Incoherence Condition does not hold, as shown in their Supplemental Information). Still their important conclusion regarding the lasso's high false discovery rate remains. 

\subsubsection*{Problem Dimension and Sparsity} 

The most obvious difference between the two studies is the scale of the problems and the true sparsity level. While Hastie~\etal considered problems with at most $10^3$ regressors, Bertsimas~\etal simulated problems up to on the order of $10^4$. Compare two problem configurations with $SNR=1$ and $(p, n, k_{true}) = (10^3, 50, 5)$ from Hastie~\etal (Fig. A.3.4) and $(10^4, 500, 50)$ from Bertsimas (Figs. 6-7), with identical linear scaling in these three quantities. At roughly similar correlation level, one would obtain comparable results in terms of variable selection (i.e. where lasso is superior to the nonconvex alternatives). It is when the amount of data available, relative to the other two quantities, is increased that the variable selection quality of the nonconvex methods significantly surpasses that of the lasso (although in the case of Hastie~et al., the relevant setting in Fig. A.4.4 $(10^3, 100, 10)$ preserves the one of the ratios by doubling $k_{true}$ with $n$, so the comparison is not exact). 

Naturally, this raises questions about the relative influence of the scaling of $(p, n, k_{true})$ on each estimator. Indeed, there is research that studies the importance of this scaling for support recovery in certain regression settings including in the case of the lasso (e.g. \cite{Wainwright2009,Wainwright2006,Gamarnik2017,Donoho2009}). In Bertsimas and Van Pary's work on CIO, they empirically demonstrate the advantage of subset selection over the lasso specifically with reference to these ``phase transitions" in variable selection \cite{Bertsimas}. Still, there is room in this area for empirical comparisons of estimators with scaling in mind, for example to put performance across problem configurations on a somewhat standardized basis and provide clarity for practitioners as to which scalings favor which estimators.

\subsection*{The Best Subset Selector: Performance and Practicality}
In step with recent advances in combinatorial optimization, (best) subset selection has gained renewed interest. There are several variations of the problem that can be solved: optimal or approximate, subset-size penalized or cardinality-constrained, regularized or unregularized, etc. Still, they are clearly members of the same family of estimators and are worth considering together.

Works from 2014-2015 by Cozad~\etal  \cite{Cozad2014} and Bertsimas~\etal \cite{Bertsimas2015} focus on using integer programming to find optimal solutions to the penalized and cardinality-constrained, respectively, versions of the best subset problem. 
However, these algorithms are computationally intensive and, as mentioned by Hastie~\etal and others \cite{Das2018}, forward stepwise selection finds very good approximate solutions at a fraction of the computational cost.

\begin{figure*}[hbt!]
\center{\includegraphics[width=\textwidth]
{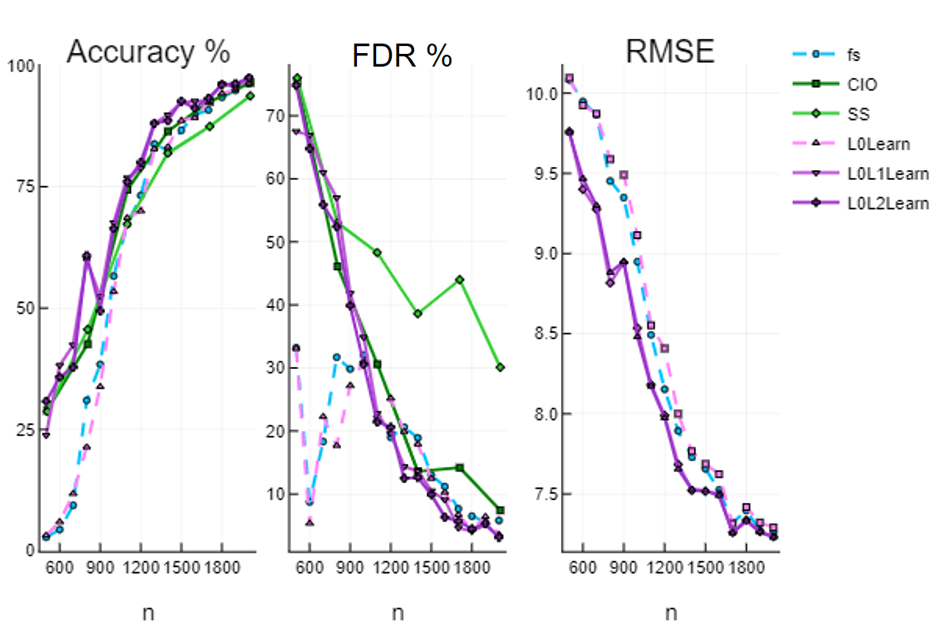}}
\caption{\label{fig:subsetcomp} Performance of various subset selection alternatives (dashed lines are used for unregularized methods). For forward-selection, the maximum number of steps was set to 100. For L0Learn, we set $(nLambda, nGamma) = (50,10)$ and used the slower but stronger CDPSI algorithm. Data for CIO/SS is taken from Bertsimas~\etal ``Accuracy \%" refers to the percentage of the true regressors identified, ``FDR \%" refers to the false discovery rate (i.e., the percentage of regressors chosen that are erroneously included), and ``RMSE" refers to the root mean square prediction error on a test set of data size $n$.}
\end{figure*}

As noted, shrinkage improves the best subset solution. As demonstrated in \textit{Spare Regression}, Bertsimas~\etal have made large algorithmic advances in optimal cardinality-constrained, $l_2$-penalized best subset selection; achieving order(s) of magnitude speedups from prior implementations. 
While tractability is a situational judgement, they argue that their CIO method is affordable for many problems. In Figure 2 of \textit{Spare Regression}, they show that the cardinality-constrained solutions for CIO/SS can be found within times one or two orders of magnitude of the lasso as implemented in the package \texttt{glmnet}. 
However, this appears to be comparing the time needed to find a single cardinality-constrained solution set to the full solution path for \texttt{glmnet}. For CIO/SS, the subset size $k$ is set to equal $k_{true}$, which results in a hard optimization problem because $p$ and $k_{true}$ are large. The value for $\gamma$, also critical for computational time, is seemingly chosen to correspond to an `appropriate' value given the noise level. However, in general application, a wide range of possibilities must be accounted for, demanding a wide range of hyperparameters be examined. In our view, finding a `reasonable'-size solution path across the two hyper-parameters $\gamma, k$ is still computationally challenging for large-scale problems.\footnote{For CIO, for example, Bertsimas~\etal note that they set a time limit of 60 s per $\gamma,k$. Assuming for the problem setting of Figures \ref{fig:rlassocomp}-\ref{fig:subsetcomp}, training is done over 100 values of $k$ and $10$ values of $\gamma$, it is reasonable to expect $> 8$ h of training time (charitably assuming that the average run time is 30 s; in reality the time limit of 60 s will be reached a majority of the time.)} Then we need to consider $K$-fold cross validation. The approximate, SS algorithm is roughly an order of magnitude faster than CIO but still comparatively quite expensive.\footnote{Please see the Supplemental Information for Bertsimas~\etal for a detailed analysis of computational time. Particularly Figure B.1.} For many cases, this cost is likely acceptable. For very large problems with thousands of regressors, many practitioners simply cannot afford it.   

Bertsimas~\etal argue that the superior performance of CIO and SS ``speaks in favor of formulations that explicitly constrain the number of features" instead of inducing sparsity via regularization. To test this claim briefly, we compare these cardinality-constrained methods with the highly-scalable L0Learn~\cite{Hazimeh2018}. L0Learn approximately solves the objective-penalized $l_0$-regularization problem with an optional $l_1$ or $l_2$ penalty (indicated as L0L1Learn/L0L2Learn). We note that the same authors, with Saab, published an algorithm called L0BnB that solves the optimal version of this $(l_0 + l_2)$-penalized problem, along with a Python-implementation\footnote{\url{https://github.com/alisaab/l0bnb}} \cite{Hazimeh2020}. We do not consider L0BnB here because the implementation is currently in the prototype stage. We use the same experimental setting\footnoteref{expnote} as in Figure \ref{fig:rlassocomp}. The results of our comparison are shown in Figure \ref{fig:subsetcomp}.

\begin{figure*}[hbt!]
\center{\includegraphics[width=\textwidth]
{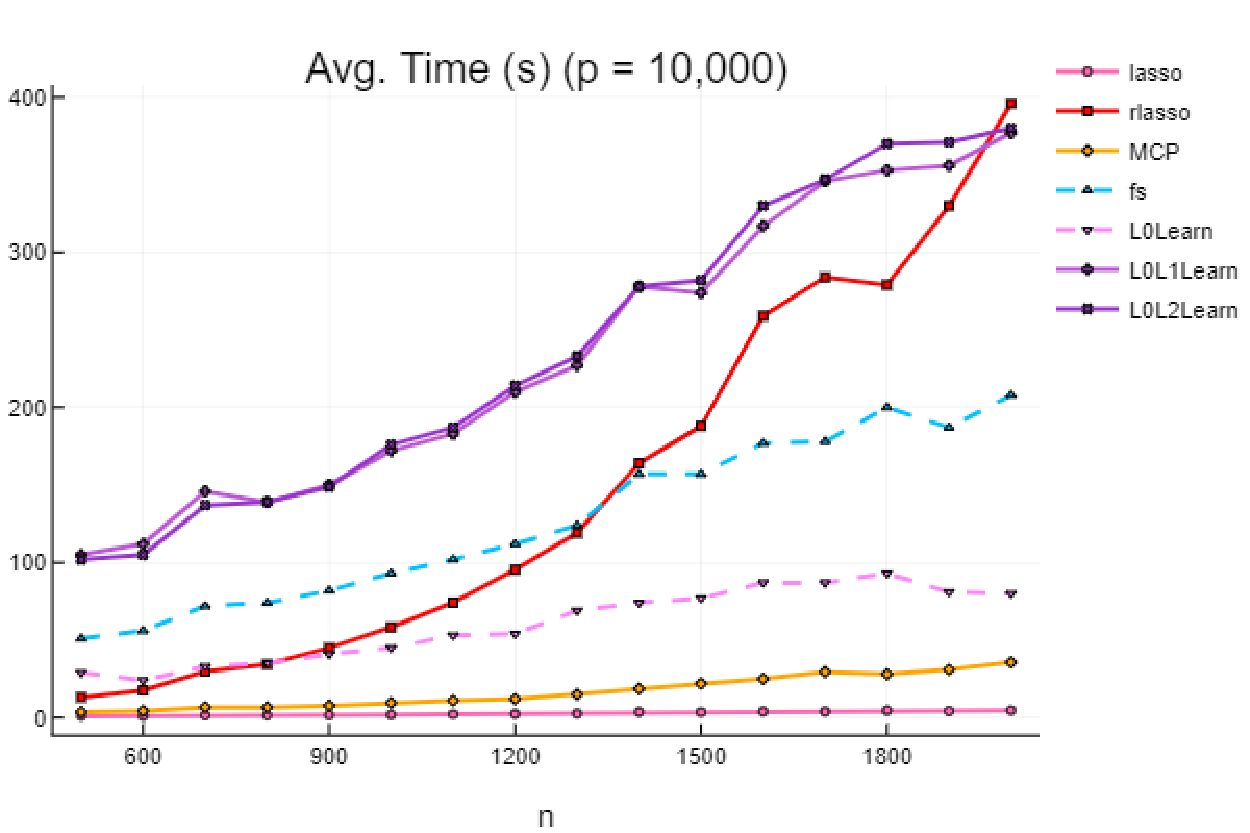}}
\caption{\label{fig:timeplot} Training time for methods considered. Time includes cross validation on separate data set of size $n$. For lasso and MCP, we used the default options from the R implementations. For rlasso, the `nlambda' option was set to 50, with 10 interpolation points each. For forward-selection, the maximum number of steps was set to 100. For L0Learn, we set $(nLambda, nGamma) = (50,10)$ and used the slower but stronger CDPSI algorithm.}
\end{figure*}

The regularized versions of the subset selection algorithms out-perform plain L0Learn and FS. This is particularly true when the amount of data is low. Once a threshold amount of data is reached, the unregularized L0Learn and FS algorithms exhibit similar performance in variable selection to their regularized counterparts. Despite this, lack of shrinkage in the coefficients ensure that test error remains larger than for the regularized L0Learn. Overall, the regularized L0Learn solutions exceed the performance of cardinality-constrained algorithms while being significantly faster and more convenient. Average time for each algorithm is plotted in Figure \ref{fig:timeplot}.

\subsection*{How do practitioners actually choose regression methods?}

It is import to understand how practitioners actually select an estimator to guide comparative research. 
In practice, convenience is the most important factor for the average modeler who is considering various methods. Accessibility of fast implementations that are also functionally labor-saving (e.g. by giving full solution paths, cross-validating, etc.) is paramount for the adoption of any estimator. If a user cannot quickly find an implementation for a new algorithm online, they will just use an established method. Although it may be best practice to test multiple algorithms or implement the intelligent combination strategy that Bertsimas~\etal suggest (where the lasso is used for dimensionality reduction, and subset selection for feature-selection), it is hard to envision a scenario where that becomes ubiquitous unless there is code that provides that functionality ``under one hood."

More sophisticated modelers may then consider problem dimensions in $(p,n)$. Although most users will lean heavily on the sparsity bet, some domain-educated modelers may have an intuition for $k_{true}$ and will consider information regarding that scaling, as well. Hastie~\etal note that the proportions of variance explained by the considered estimators do not differ much, meaning that it ``makes sense overall to favor the methods that are easy to compute.'' However, Bertsimas~\etal would likely justifiably argue that, even if there is no obvious reason to assume \textit{a priori} that one estimator would have a higher PVE (or lower test error) than another, those who value model conciseness may prefer $l_0$-methods when the scaling seems appropriate. 

Even more advanced users may have some expectations of sparsity structure, correlation structure, and noise that will guide them towards particular, or even specialized, estimators. Users may also desire to account for constraints on the response surface, which model-building algorithms like \cite{Cozad2015} can enforce explicitly. 

\section*{Conclusions}
Hastie~\etal emphasize that they are \textit{not} trying to determine the ``best variable selector" or identify the ``best prediction algorithm." Indeed, even if it was possible to define and agree upon what qualities make a regression algorithm ``best", in a noisy and nonlinear world, it is impossible to speak in absolutes. 

But if the ultimate goal of statistics is to create value, then even if we cannot pin-down the idea of ``best", we must certainly be able to point in its general direction. Both Bertsimas~\etal and Hastie~\etal inexorably point towards methods that they think are favorable in particular settings. Their work taken together provides useful intuition about when to apply which methods, and where we can improve and learn more. 

So, let us again step into the mind of an average user. Given only a data set, and assuming that we value convenience, how should we go-about building a model?

If our priority is predictive or explanatory performance, it seems most natural to ``favor the methods that are easy to compute" and start with the lasso/ENet using the \texttt{glmnet}\footnote{\href{https://cran.r-project.org/web/packages/glmnet/index.html}{R Package Link}|\href{https://github.com/bbalasub1/glmnet_python}{Python}| \href{https://web.stanford.edu/~hastie/glmnet_matlab/download.html}{MATLAB}} package, which is very powerful, fast, and available in multiple langauges. It would also make sense to benchmark against a sparser estimator, for which MCP (via \texttt{ncvreg}\footnoteref{ncvregnote} \cite{Breheny2011}) is a trusted choice that is efficient and convenient. Suppose that the MCP solution is appreciably sparser than the model from \texttt{glmnet}. Then, even if this sparser model does not lower test error, we might still prefer a more interpretable solution. To this end, rlasso\footnoteref{bestsubsetref} shows robust performance and sparsity. We can expect more sparsity with a modest increase in computational time\footnote{\label{timenote}Importantly, Hazimeh~\etal \cite{Hazimeh2018} report that the faster `Algorithm 1' achieved notable speedups of 25-300\% over \texttt{glmnet} and \texttt{ncvreg} for very large instances and performed comparatively well on real problems. We did not test this algorithm because we wanted to maximize the regression performance metrics and, in \cite{Hazimeh2018}, the more intensive CDPSI algorithm performed substantially better than Algorithm 1 on synthetic data. Still, we acknowledge that \texttt{L0Learn} with the (default) Algorithm 1 may be the most convenient option.} by performing regularized subset selection using L0Learn as implemented in the \texttt{L0Learn}\footnoteref{l0learnref} package, configured with the added $l_1/l_2$-penalty and solved using the stronger CDPSI algorithmic option.

If our priority is sparsity, as Bertsimas~\etal argue, nonconvex methods achieve just that without necessarily sacrificing predictive power. As mentioned, MCP and rlasso are fast and accessible, but subset-selection via regularized \texttt{L0Learn} will provide users (who can afford the potential extra cost \footnoteref{timenote}) confidence that they are approaching the current limits of achievable sparsity. In the most general case, we would not turn to the cardinality-constrained estimators CIO/SS, as implemented in the \texttt{SubsetSelectionCIO.jl}\footnoteref{cioref} and \texttt{SubsetSelection.jl}\footnoteref{ssref} packages, because of the relatively demanding computational time needed and lack of seamless hyperparameter selection and tuning that methods like \texttt{L0Learn} provide (by calculating hyperparameter-paths from the data). We would also avoid unregularized subset-selection algorithms like forward-selection, BSS, or the corresponding configuration in \texttt{L0Learn} because of their poor performance on noisy data.  

It is impossible to give broad advice for all conceivable situations and the experiments in this paper and the two discussed can only capture a portion of the range of problem configurations. There is still progress to be made in translating experimental results into specific prescriptions. We invite readers to expand upon the ideas presented in the studies highlighted here and share their results as part of a continuing discussion. 

\section*{Acknowledgment}
This work was conducted as part of the Institute for the Design of Advanced Energy Systems (IDAES) with funding from the Office of Fossil Energy, Cross-Cutting Research, U.S. Department of Energy.

\bibliographystyle{plain}
\bibliography{library.bib} 

\begin{thebibliography}{10}

\bibitem{Bertsimas2015}
D.~Bertsimas, A.~King, and R.~Mazumder.
\newblock {Best Subset Selection via a Modern Optimization Lens}.
\newblock {\em The Annals of Statistics}, 44(2):813--852, 2015.

\bibitem{Bertsimas}
D.~Bertsimas and B.~V. Parys.
\newblock {Sparse High-Dimensional Regression: Exact Scalable Algorithms and
  Phase Transitions}.
\newblock {\em Annals of Statistics}, 48(1), 2020.

\bibitem{Breheny2011}
P.~Breheny and J.~Huang.
\newblock {Coordinate descent algorithms for nonconvex penalized regression,
  with applications to biological feature selection}.
\newblock {\em The Annals of Applied Statistics}, 5(1):232--253, 2011.

\bibitem{Breiman1995}
L.~Breiman.
\newblock {Better subset regression using the Nonnegative Garrote}.
\newblock {\em Technometrics}, 37(4):373--384, 1995.

\bibitem{Candes2007}
E.~Candes and T.~Tao.
\newblock {The Dantzing selector: Statistical estimation when $p$ is much
  larger than $n$}.
\newblock {\em The Annals of Statistics}, 35(6):2313--2351, 2007.

\bibitem{Celentano2019}
M.~Celentano and A.~Montanari.
\newblock {Fundamental Barriers to High-Dimensional Regression with Convex
  Penalties}.
\newblock {\em arXiv:1903.10603}, 2019.

\bibitem{Cozad2014}
A.~Cozad, N.~V. Sahinidis, and D.~C. Miller.
\newblock {Learning surrogate models for simulation-based optimization}.
\newblock {\em AIChE Journal}, 60(6):2211--2227, 2014.

\bibitem{Cozad2015}
A.~Cozad, N.~V. Sahinidis, and D.~C. Miller.
\newblock {A combined first-principles and data-driven approach to model
  building}.
\newblock {\em Computers {\&} Chemical Engineering}, 73:116--127, 2015.

\bibitem{Das2018}
A.~Das and D.~Kempe.
\newblock {Approximate Submodularity and its Applications: Subset Selection,
  Sparse Approximation and Dictionary Selection}.
\newblock {\em Journal of Machine Learning Research}, 19:1--34, 2018.

\bibitem{Donoho2009}
D.~L. Donoho and J.~Tanner.
\newblock {Observed universality of phase transitions in high-dimenstional
  geometry, with implications for modern data analysis and signal processing}.
\newblock {\em Philosophical Transactions of the Royal Society B: Biological
  Sciences}, 367(1906), 2009.

\bibitem{Fan2001}
J.~Fan and R.~Li.
\newblock {Variable selection via nonconcave penalized likelihood and its
  oracle properties}.
\newblock {\em Journal of the American Statistical Association},
  96(456):1348--1360, 2001.

\bibitem{Frank1993}
I.~E. Frank and J.~H. Friedman.
\newblock {A Statistical View of Some Chemometrics Regression Tools}.
\newblock {\em Technometrics}, 35(2):109--135, 1993.

\bibitem{Furnival1974}
G.~M. Furnival and R.~W. Wilson.
\newblock {Regressions by Leaps and Bounds}.
\newblock {\em Technometrics}, 16(4):499--511, 1974.

\bibitem{Gamarnik2017}
D.~Gamarnik and I.~Zadik.
\newblock High-dimensional regression with binary coefficients. estimating
  squared error and a phase transition.
\newblock {\em arXiv:1701.04455}, 2017.

\bibitem{Hastie2015}
T.~Hastie, R.~Tibshirani, and R.~Wainwright.
\newblock {\em {Statistical Learning with Sparsity: The Lasso and
  Generalizations}}.
\newblock CRC Press, 2015.

\bibitem{Hazimeh2018}
H.~Hazimeh and R.~Mazumder.
\newblock Fast best subset selection: Coordinate descent and local
  combinatorial optimization algorithms.
\newblock {\em arXiv:1803.01454}, 2018.

\bibitem{Hazimeh2020}
H.~Hazimeh, R.~Mazumder, and A.~Saab.
\newblock Sparse regression at scale: Branch-and-bound rooted in first-order
  optimization.
\newblock {\em arXiv:2004:06152}, 2020.

\bibitem{Hocking1967}
R.~R. Hocking and R.~N. Leslie.
\newblock {Selection of the Best Subset in Regression Analysis}.
\newblock {\em Technometrics}, 9(4):531, 1967.

\bibitem{Hoerl1970}
A.~E. Hoerl and R.~W. Kennard.
\newblock {Ridge regression: Biased estimation for nonorthogonal problems}.
\newblock {\em Technometrics}, 12(1):55--67, 1970.

\bibitem{Loh2017}
P.~Loh and M.~J. Wainwright.
\newblock {Support recovery without incoherence: A case for nonconvex
  regularization}.
\newblock {\em The Annals of Statistics}, 45(6):2455--2482, 2017.

\bibitem{Mazumder2011}
R.~Mazumder, J.~H. Friedman, and T.~Hastie.
\newblock Sparsenet: Coordinate descent with nonconvex penalties.
\newblock {\em Journal of the American Statistical Association},
  106(495):1125--1138, 2011.
\newblock PMID: 25580042.

\bibitem{Meinshausen2007}
N.~Meinshausen.
\newblock {Relaxed Lasso}.
\newblock {\em Computational Statistics {\&} Data Analysis}, 52(1):374--393,
  2007.

\bibitem{Meinshausen2010}
N.~Meinshausen and P.~B{\"{u}}hlmann.
\newblock {Stability selection}.
\newblock {\em Journal of the Royal Statistical Society: Series B (Statistical
  Methodology)}, 72(4):417--473, 2010.

\bibitem{Pilanci2015}
M.~Pilanci, M.~J.{\textperiodcentered} Wainwright, and L.~{El Ghaoui}.
\newblock {Sparse learning via Boolean relaxations}.
\newblock {\em Math. Program., Ser. B}, 151:63--87, 2015.

\bibitem{Stigler1988}
S.~M. Stigler.
\newblock {Gauss and the invention of least squares}.
\newblock {\em Annals of Statistics}, 2(5):347--370, 1988.

\bibitem{Tibshirani1996}
R.~Tibshirani.
\newblock {Regression Selection and Shrinkage via the Lasso}.
\newblock {\em Journal of the Royal Statistical Society}, 58(1):267--288, 1996.

\bibitem{Wainwright2009}
M.~J. Wainwright.
\newblock {Information-theoretic limits on sparsity recovery in the
  high-dimensional and noisy setting}.
\newblock {\em IEEE Transactions on Information Theory}, 55(12):5728--5741,
  2009.

\bibitem{Wainwright2006}
M.~J. Wainwright.
\newblock {Sharp thresholds for high-dimensional and noisy recovery of
  sparsity}.
\newblock {\em IEEE Transactions on Information Theory}, 55(5):2183--2202,
  2009.

\bibitem{Wang2020}
F.~Wang, S.~Mukherjee, S.~Richardson, and S.~M. Hill.
\newblock {High-dimensional regression in practice: an empirical study of
  finite-sample prediction, variable selection and ranking}.
\newblock {\em Statistics and Computing}, 30:697--719, 2020.

\bibitem{Wang2011}
S.~Wang, B.~Nan, S.~Rosset, and J.~Zhu.
\newblock {Random lasso}.
\newblock {\em The Annals of Applied Statistics}, 5(1):468--485, 2011.

\bibitem{Zhang2010}
C.~Zhang.
\newblock {Nearly unbiased variable selection under minimax concave penalty}.
\newblock {\em The Annals of Statistics}, 38(2):894--942, 2010.

\bibitem{Zou2006}
H.~Zou.
\newblock {The Adaptive Lasso and its oracle properties}.
\newblock {\em Journal of the American Statistical Association},
  101(476):1418--1429, 2006.

\bibitem{Zou2005}
H.~Zou and T.~Hastie.
\newblock {Regularization and variable selection via the elastic net}.
\newblock {\em Journal of the Royal Statistical Society: Series B (Statistical
  Methodology)}, 67(2):301--320, 2005.

\end{thebibliography}
\end{singlespacing}

\end{document}